\documentclass{bmvc2k}

\usepackage{multirow}
\usepackage{amssymb}
\usepackage{booktabs}
\usepackage{appendix}
\usepackage{hyperref}

\hypersetup{hidelinks,
	colorlinks=true,
	allcolors=blue,
	pdfstartview=Fit,
	breaklinks=true}

\title{Local Implicit Wavelet Transformer for Arbitrary-Scale Super-Resolution}

\addauthor{Minghong Duan}{22111010025@m.fudan.edu.cn}{1,2}
\addauthor{Linhao Qu}{lhqu20@fudan.edu.cn}{1,2}
\addauthor{Shaolei Liu}{19111010029@fudan.edu.cn}{1,2}
\addauthor{Manning Wang}{mnwang@fudan.edu.cn}{1,2*}

\addinstitution{
 Digital Medical Research Center\\
 Fudan University\\
 Shanghai, China
}
\addinstitution{
Shanghai Key Lab of Medical\\ Image Computing and Computer\\ Assisted Intervention
}

\runninghead{DUAN ET AT.}{Local Implicit Wavelet Transformer}

\begin{document}

\maketitle

\begin{abstract}
Implicit neural representations have recently demonstrated promising potential in arbitrary-scale Super-Resolution (SR) of images. Most existing methods predict the pixel in the SR image based on the queried coordinate and ensemble nearby features, overlooking the importance of incorporating high-frequency prior information in images, which results in limited performance in reconstructing high-frequency texture details in images. To address this issue, we propose the Local Implicit Wavelet Transformer (LIWT) to enhance the restoration of high-frequency texture details. Specifically, we decompose the features extracted by an encoder into four sub-bands containing different frequency information using Discrete Wavelet Transform (DWT). We then introduce the Wavelet Enhanced Residual Module (WERM) to transform these four sub-bands into high-frequency priors, followed by utilizing the Wavelet Mutual Projected Fusion (WMPF) and the Wavelet-aware Implicit Attention (WIA) to fully exploit the high-frequency prior information for recovering high-frequency details in images. We conducted extensive experiments on benchmark datasets to validate the effectiveness of LIWT. Both qualitative and quantitative results demonstrate that LIWT achieves promising performance in arbitrary-scale SR tasks, outperforming other state-of-the-art methods. The code is available at \href{https://github.com/dmhdmhdmh/LIWT}{https://github.com/dmhdmhdmh/LIWT}.
\end{abstract}

\section{Introduction}
\label{sec:intro}
\begin{figure}[t]
\setlength{\abovecaptionskip}{-0.5cm}
\setlength{\belowcaptionskip}{-0.5cm}
\centerline{\includegraphics[width=\columnwidth]{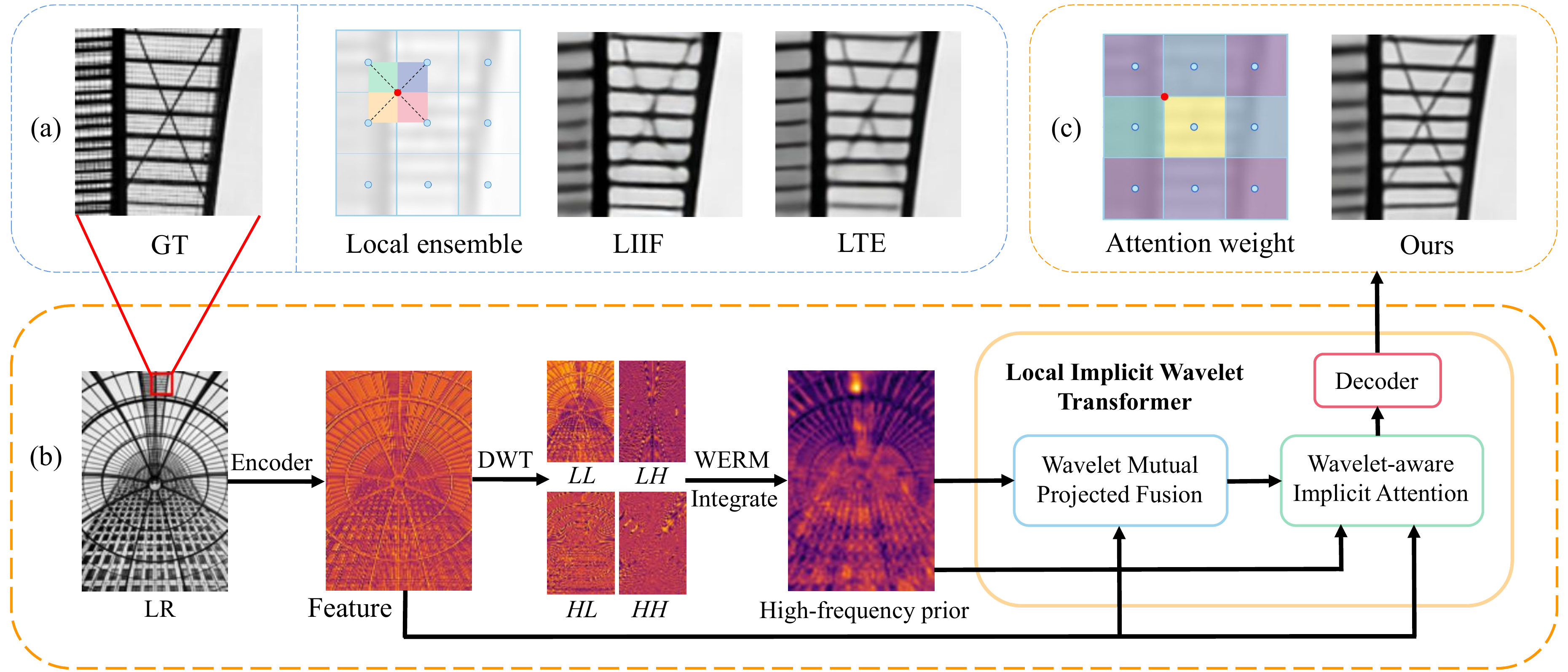}}
\caption{Motivation and effectiveness of our method. (a) LIIF~\cite{chen2021learning} and LTE~\cite{lee2022local} have difficulty reconstructing high-frequency details using the local ensemble technique. (b) LIWT introduces the high-frequency prior via DWT. (c) LIWT can reconstruct high-frequency details using attention weight based on the high-frequency prior.}
\label{fig1}
\end{figure}

Single Image Super-Resolution (SISR) refers to the process of recovering a high-resolution (HR) image from a single low-resolution (LR) image and has been widely applied across various fields \cite{shi2013cardiac, thornton2006sub, zou2011very, gunturk2004super, duan2024towards}. Most existing SISR models comprise a deep neural network (DNN) with an upsampling module like learnable deconvolutions or pixel shuffling \cite{dong2016accelerating, shi2016real} and can only deal with integer scaling factors, and these models necessitate retraining when encountering new scaling factors. Recent work has achieved arbitrary-scale super-resolution (SR) by replacing the upsampling layer typically used in previous methods with a local implicit image function and has demonstrated exemplary performance \cite{chen2021learning,lee2022local}. These methods based on local implicit functions first extract features from LR images through a DNN-based encoder and then employ multi-layer perceptrons (MLPs) to map the 2D query coordinates of the HR image and the aggregated representation of the corresponding local region features (called latent code) to RGB values. There are two limitations to these existing methods. Firstly, the coordinate-based local ensemble technique \cite{chen2021learning,lee2022local} used for querying RGB values fails to consider the relevance of features within local regions. Ensemble weights are typically computed based on the rectangular area between the query point and each nearest point (Figure \ref{fig1}(a)). These weights are solely dependent on the positional relationship between the query point and its nearest coordinates of local features and do not account for the features themselves, thus limiting the reconstruction performance of the model. Secondly, only the four nearest latent codes to the query point are used when querying RGB values based on coordinates. We argue that the representational capacity of LR features directly obtained from the encoder is limited, especially in large-scale SR, which may lead to blurry results lacking texture details. Introducing high-frequency prior information of image features into the local implicit functions is therefore necessary.\\
\indent Many existing methods have shown that high-frequency prior information obtained from discrete wavelet transform (DWT) can improve the performance of SR models based on deep learning \cite{hsu2023wavelet,zou2022joint,xin2020wavelet}. However, the scale transformation rate of the DWT when applied to features or images is limited to powers of 2, and most DWT-based methods rely on this property for inverse transformation to achieve upsampling, thereby not achieving arbitrary-scale SR. To better restore high-frequency details while achieving arbitrary resolution upscaling, we propose the Local Implicit Wavelet Transformer (LIWT), which leverages cross-attention to exploit the high-frequency information obtained from DWT fully and accounts for the relevance of the features within a local region. As shown in Figure \ref{fig1}(b), LIWT consists of a Wavelet Mutual Projected Fusion (WMPF), a Wavelet-aware Implicit Attention (WIA), and a decoder. We first extract features from the LR image using an encoder and then decompose the features obtained into low-frequency components $LL$ and high-frequency components $LH$, $HL$, and $HH$ using DWT. To enhance LIWT's capability to capture high-frequency details, we designed the Wavelet Enhancement Residual Module (WERM) to integrate the four components obtained by DWT and output features with high-frequency priors. Subsequently, we fuse the high-frequency priors with the features from the encoder using WMPF to assist in reconstruction. Then, we employed WIA to generate attention maps based on query coordinates and sample nearest-neighbor interpolated latent vectors from both the high-frequency prior features and the original features. By applying these attention maps to these feature embeddings, LIWT focuses more on high-frequency details in the image. Finally, the decoder utilizes attention feature embeddings to generate RGB values. As illustrated in Figure \ref{fig1}(c), employing attention weight based on high-frequency priors enables the reconstruction of high-frequency details.\\
\indent The main contributions of our work are summarized as follows: (1) We introduce the Local Implicit Wavelet Transformer (LIWT), which integrates features obtained from DWT into local implicit image functions using the designed WERM and WIA to enhance performance; (2) We demonstrate that LIWT can be effectively integrated into different encoders to enhance performance, outperforming other arbitrary-scale SR methods; (3) We conduct a comprehensive analysis of LIWT. Extensive experimental results demonstrate that the proposed LIWT can produce superior or comparable results on benchmark datasets.
\vspace{-1.0em}
\section{Related work\vspace{-0.2em}}
{\bf Single image super-resolution.} SRCNN~\cite{dong2015image} pioneered the implementation of SISR using a convolutional neural network (CNN) in an end-to-end manner, subsequently leading to a series of methods that utilized other CNN modules designed to enhance SR performance further. Examples include methods employing residual modules such as EDSR~\cite{lim2017enhanced} or dense connection modules such as RDN~\cite{zhang2018residual}. In addition, methods based on attention mechanisms \cite{zhang2018image,niu2020single,dai2019second,liang2021swinir,chen2021pre,mei2021image,xia2022efficient} have been introduced into SISR, including channel attention \cite{zhang2018image,niu2020single,dai2019second}, self-attention \cite{liang2021swinir,chen2021pre}, and non-local attention \cite{mei2021image,xia2022efficient}. \\
{\bf DWT based image super-resolution.} DWT has been widely used in SISR due to its ability to express different frequency components \cite{liu2019multi,guo2017deep,xue2020wavelet,hsu2023wavelet,zou2022joint,xin2020wavelet}. To leverage the high-frequency representation capability of DWT, WDRN~\cite{xin2020wavelet} guides the feature extraction process to preserve high-frequency features in the wavelet domain, reconstructing HR images with more precise details; JWSGN~\cite{zou2022joint} utilizes DWT to transform input features into the frequency domain and generate edge feature maps to correct high-frequency components, further recovering high-frequency details. Despite their impressive results, these SISR models require training different models for each upscaling factor, limiting their applicability.\\
{\bf Arbitrary-scale super-resolution.} Several methods have been proposed to train a unified model capable of handling arbitrary upscaling factors \cite{chen2021learning,lee2022local,hu2019meta,xu2021ultrasr,liu2021enhancing,chen2023cascaded}. LIIF~\cite{chen2021learning} employs an MLP as a local implicit function, which predicts the RGB values at any query coordinate by acquiring the HR image coordinates and the features around the coordinates. LTE~\cite{lee2022local} further introduces a local texture estimator that enriches the representation capability of local implicit functions by predicting Fourier information. Unlike the above methods, we introduce the high-frequency prior obtained from the wavelet transform into the local implicit function, further improving the recovery of high-frequency details.\\
\vspace{-1.2em}
\begin{figure}[h]
\setlength{\abovecaptionskip}{-0.5cm}
\setlength{\belowcaptionskip}{-0.5cm}
\centerline{\includegraphics[width=\columnwidth]{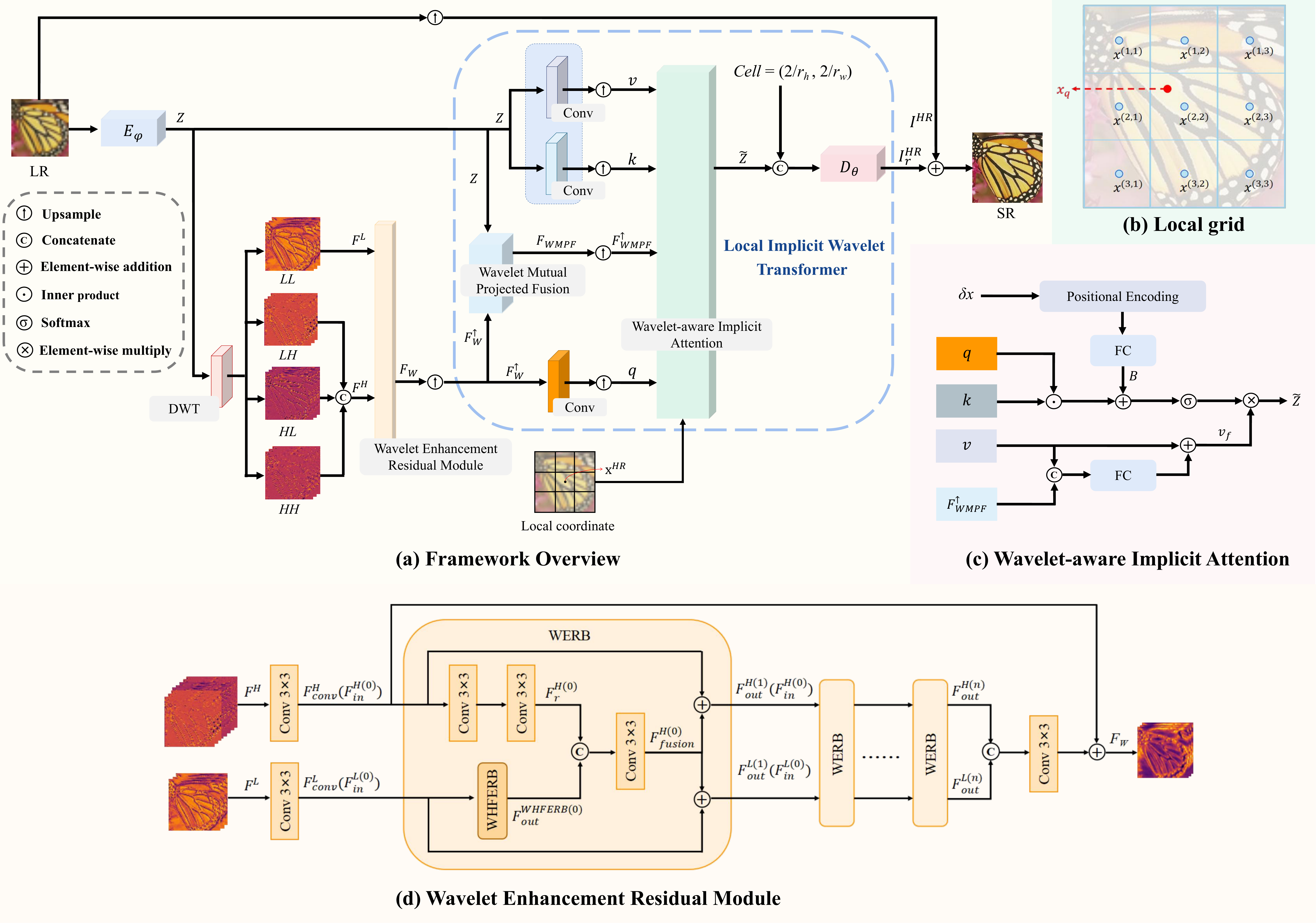}}
\caption{(a) Overview of the proposed framework. (b) Diagram of the local grid. (c) Structure of the WIA. (d) Structure of the WERM.}
\label{fig2}
\end{figure}
\vspace{-1.0em}
\section{Methodology \vspace{-0.2em}}
The overall framework of the proposed model, which integrates the encoder, the Wavelet Enhancement Residual Module (WERM), and the LIWT, is given in Figure \ref{fig2}(a). For a given LR image $I^{LR}\in R^ {H\times W\times 3}$ at the 2D LR coordinates $x^{LR} \in \chi$ and any magnification $s=\left\{s_h, s_w\right\}$, the model outputs an HR image $I^{HR}\in R^ {s_hH\times s_wW\times 3}$ at the 2D HR coordinates $x^{HR} \in \chi$, where $\chi$ is the 2D coordinate space used to express the continuous 2D image domain. First, we employ the encoder $E_\varphi$ to extract features from the LR image, obtaining $Z\in R^ {H\times W\times C}$. Subsequently, we decompose $Z$ using discrete wavelet transform (DWT) into low-frequency component \emph{LL} and high-frequency components \emph{LH}, \emph{HL}, and \emph{HH}. The low-frequency component \emph{LL} is denoted as $F^L$, while the high-frequency components \emph{LH}, \emph{HL}, and \emph{HH} are concatenated along the channel axis and represented as $F^H$. We further process $F^L$ and $F^H$ using WERM to isolate high-frequency information, yielding enhanced texture features $F_W\in R^ {H/2\times W/2\times C}$ as high-frequency priors. We perform bicubic upsampling of $F_W$ to obtain $F_W^\uparrow \in R^ {H\times W\times C}$ and input it with the feature $Z$ and the 2D HR coordinate $x^{HR}$ into the LIWT to generate the RGB values of the residual image $I_r^{HR}\in R^ {s_hH\times s_wW\times 3}$. Finally, the residual image $I_r^{HR}\in R^ {s_hH\times s_wW\times 3}$ is summed via element-wise addition with the bilinearly upsampled image $I_\uparrow^{HR}\in R^ {s_hH\times s_wW\times 3}$ to generate the final predicted HR image $I^{HR}$. LIWT consists of the Wavelet Mutual Projected Fusion (WMPF), the Wavelet-aware Implicit Attention (WIA), and a decoder, collectively utilizing the high-frequency prior $F_W$ to recover image details. We train the entire framework using the L1 loss and utilize the Haar wavelet transform to perform the DWT.
\vspace{-1.0em}
\subsection{Wavelet Enhancement Residual Module}
\begin{figure}[h]
\centerline{\includegraphics[width=\columnwidth]{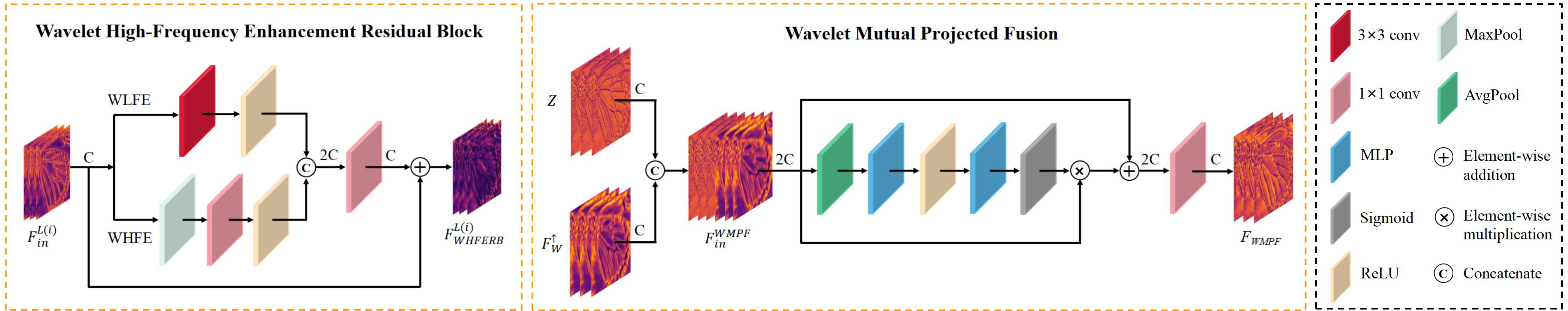}}
\caption{Structure of the WHFERB and the WMPF.}
\label{fig3}
\end{figure}

To extract more high-frequency information conducive to reconstruction from the different components obtained through DWT decomposition, we devised WERM to process $F^L$ and $F^H$, which comprises several wavelet enhancement residual blocks (WERBs). As illustrated in Figure \ref{fig2}(d), the input to the kth WERB is denoted as $F_{in}^{L(k)}$ and $F_{in}^{H(k)}$, and the output is denoted as $F_{out}^{L(k)}$ and $F_{out}^{H(k)}$. Initially, we preprocess $F^L$ and $F^H$ separately using a $3\times3$ convolutional layer to obtain $F_{conv}^{L}$ and $F_{conv}^{H}$, which are then fed into the first WERB. We further designed a wavelet high-frequency enhancement residual block (WHFERB) to separate more sharp edge components from the smoothed surface of $F_{in}^{L(k)}$ and assist in reconstruction at each WERB. Figure \ref{fig3} shows WHFERB consists of a Wavelet Local Feature Extraction (WLFE) branch and a Wavelet High-Frequency Enhancement (WHFE) branch. For the WLFE branch, we employ a $3\times3$ convolutional layer followed by a ReLU activation function to extract local features. For the WHFE branch, we employ a max-pooling layer to extract high-frequency information from the input features and use a $3\times3$ convolutional layer followed by a ReLU activation function to enhance the high-frequency features. Subsequently, we concatenate the outputs of the two branches and input them into a $1\times1$ convolutional layer for information fusion. We refine the high-frequency feature $F_{in}^{H(k)}$ through two consecutive $3\times3$ convolutional layers to obtain $F_r^{H(k)}$. Then, the outputs of the two branches are concatenated and input into a $3\times3$ convolutional layer for information fusion to obtain $F_{fusion}^{H(k)}$. To effectively utilize high-frequency information and maintain training stability, we introduce skip-connections to transmit the fused information $F_{fusion}^{H(k)}$ back to the two branches separately, allowing the frequency information of the two branches to complement each other. For the WERM containing n WERBs, we concatenate the outputs $F_{out}^{L(n)}$ and $F_{out}^{H(n)}$ of the nth WERB and fuse them using a $3\times3$ convolutional layer. We also introduce skip connections to add the fused result to $F_{conv}^H$ to obtain the final high-frequency prior feature representation $F_W$. In this paper, we set n to 4.

\vspace{-1.2em}

\subsection{Local Implicit Wavelet Transformer}
{\bf Overview.} LIWT processes the high-frequency prior feature $F_W$ and features Z from the encoder, and it outputs the RGB values of the corresponding residual image $I_r^{HR}\in R^ {s_hH\times s_wW\times 3}$ based on the input 2D HR coordinates $\mathbf{x}^{HR}$. Figure \ref{fig2}(a) illustrates that it consists of the WMPF, the WIA, and a decoder $D_\theta$ parameterized by $\theta$. To leverage the useful texture information from the high-frequency prior $F_W$, we perform bicubic upsampling on  $F_W$ to obtain $F_W^\uparrow$ firstly, which has the same size as feature $Z$. Subsequently, we assign weight coefficients to each channel of $F_W^\uparrow$ and feature $Z$ and then fuse them to yield $F_{WMPF}$ through WMPF. This operation aims to direct the LIWT to focus more on the high-frequency components in $F_W^\uparrow$ that are beneficial for detail recovery. Simultaneously, we use convolutional layers to project $F_W^\uparrow$ and derive the latent embedding corresponding to the query $q$. Additionally, using two separate convolutional layers, we project $Z$ to obtain latent embeddings for key $k$ and value $v$, respectively. WIA estimates the local latent embedding $\tilde{z} \in {R}^{9 \times C}$ for HR coordinates within ${3\times3}$ local grids based on the input $q$, $k$, $v$, and $F_{WMPF}$. Finally, $D_\theta$ utilizes these embeddings and the provided cell to predict the RGB values $I_r^{HR}$ at the HR coordinates. The process of LIWT can be formalized as follows: 
\begin{equation}
F_{WMPF}=WMPF\left(F_W^{\uparrow}, Z\right)
\end{equation}
\begin{equation}
\tilde{z}=W I A\left(\delta \mathbf{x}, F_{W M P F}^{\uparrow}, q, k, v\right),  \delta \mathbf{x}=\left\{x_q-x^{(i, j)}\right\}_{i \in\{1,2,3\}, j \in\{1,2,3\}}
\end{equation}
\begin{equation}
I^r\left(x_q\right)=D_\theta(\tilde{z}, c)
\end{equation}
where $x^{(i, j)}$ represents the LR coordinates within the local grids that are closest to the HR coordinates $x_q$, as shown in Figure \ref{fig2}(b). The query latent code $q \in {R}^{1 \times C}$ at the HR coordinate $x_q$ is calculated by nearest neighbor interpolation, and the feature embeddings $F_{WMPF}^{\uparrow} \in {R}^{9 \times C}$, $k \in {R}^{9 \times C}$, and $v \in {R}^{9 \times C}$ are sampled at the local LR coordinate $\mathbf{x}=\left\{x^{i, j}\right\}_{i \in\{1,2,3\}, j \in\{1,2,3\}}$. $I^r\left(x_q\right)$ represents the predicted RGB value at the query coordinate $x_q$, $c=\left\{2 / s_h H, 2 / s_w W\right\}$ denotes the unit cell representing the height and width of pixels in the HR image, and $D_\theta$ is implemented as a five-layer MLP.\\
{\bf Wavelet Mutual Projected Fusion.} LIWT utilizes WMPF to fuse the high-frequency details in $F_W^\uparrow$ with the features $Z$ from the encoder. As illustrated in Figure \ref{fig3}, WMPF first concatenates $F_W^\uparrow$ and feature $Z$ along the channel dimension to form the input $F_{in}^{WMPF}$. Subsequently, the entire input is represented through global average pooling, followed by learning the non-linear interactions within the concatenated features $F_{in}^{WMPF}$ across channels using $FC$-$ReLU$-$FC$ layers, where $FC$ represents a fully connected layer. Finally, attention coefficients are obtained via a sigmoid activation function, which is then multiplied with the features $F_{in}^{WMPF}$ at the channel level, enabling effective focus on the high-frequency components and contextual information of the features. Additionally, skip-connections are introduced to maintain training stability, and the fused features $F_{WMPF}$ are outputted through a $1\times1$ convolutional layer.\\
{\bf Wavelet-aware Implicit Attention.} As shown in Figure \ref{fig2}(c), LIWT utilizes WIA to perform cross-attention on the local grid, generating local latent embeddings $\tilde{z} \in {R}^{9 \times C}$ for each HR coordinate. To effectively utilize the fused features $F_{WMPF}$ in the attention mechanism and avoid generating incorrect textures, we further fuse $F_{WMPF}^{\uparrow}$ and value $v$ to obtain $v_f$:
\begin{equation}
v_f=F C\left(\operatorname{Concat}\left(v, F_{W M P F}^{\uparrow}\right)\right)+v
\end{equation}
We define $q \in {R}^{1 \times C}$, $k \in {R}^{9 \times C}$, and $v \in {R}^{9 \times C}$ on the local grid as sets $q=\left\{q_\tau \mid \tau=1,2, \ldots, H\right\}$, $k=\left\{k_\tau^r \mid r=1,2, \ldots, 9, \tau=1,2, \ldots, H\right\}$, and $v_f=\left\{v_\tau^r \mid r=1,2, \ldots, 9, \tau=1,2, \ldots, H\right\}$, where $r$ represents the latent code index on local grid, $H$ represents the number of attention heads, and we set $H$ to 8. WIA first calculates the inner product of $q$ and $k$, then adds the relative position deviation $B$ to the calculation result to obtain the attention matrix. Subsequently, the attention matrix is normalized through Softmax operation to generate a local attention map. Finally, WIA performs element-wise multiplication on $v_f$ and the local attention map and concatenates the operation results of different attention heads to obtain $\tilde{z}$:
\begin{equation}
\tilde{z}=\operatorname{Concat}\left(\left\{\frac{\exp \left(\frac{q_\tau^r\left(k_\tau^r\right)^{\top}}{\sqrt{C / H}}+B\right)}{\sum_{r=1}^9 \exp \left(\frac{q_\tau^r\left(k_\tau^r\right)^{\top}}{\sqrt{C / H}}+B\right)} \times v_\tau^r\right\}_{\tau=1,2, \ldots, H}\right)
\end{equation}
\begin{equation}
B=F C(\gamma(\delta \mathbf{x})),  \gamma(\delta \mathbf{x})=\left[\sin \left(2^0 \delta \mathbf{x}\right), \cos \left(2^0 \delta \mathbf{x}\right), \ldots, \sin \left(2^{L-1} \delta \mathbf{x}\right), \cos \left(2^{L-1} \delta \mathbf{x}\right)\right]
\end{equation}
where $C$ represents the dimension of the latent embedding, $\gamma$ is the position encoding function, and $L$ is the encoding dimension hyperparameter and is set to 10.

\begin{table}[h]
\setlength{\abovecaptionskip}{-0.5cm}
\setlength{\belowcaptionskip}{-0.5cm}
\begin{center}
\resizebox{0.77\columnwidth}{!}{
\begin{tabular}{c|c|cccccccc}
\hline
Backbone                 & Methods            & $\times2$                           & $\times3$                           & $\times4$                           & $\times6$                           & $\times12$                          & $\times18$                          & $\times24$                          & $\times30$                          \\ \hline
-                        & Bicubic            & 31.01                        & 28.22                        & 26.66                        & 24.82                        & 22.27                        & 21                           & 20.19                        & 19.59                        \\ \hline
                         & EDSR-baseline~\cite{lim2017enhanced}      & 34.55                        & 30.9                         & 28.94                        & -                            & -                            & -                            & -                            & -                            \\
                         & EDSR-Meta-SR~\cite{hu2019meta}
                         & 34.64                        & 30.93                        & 28.92                        & 26.61                        & 23.55                        & 22.03                        & 21.06                        & 20.37                        \\
                         & EDSR-LIIF~\cite{chen2021learning}         & 34.67                        & 30.96                        & 29.00                        & 26.75                        & 23.71                        & 22.17                        & 21.18                        & 20.48                        \\
                         & EDSR-UltraSR~\cite{xu2021ultrasr}       & 34.69                        & {\color[HTML]{0000FF} 31.02} & {\color[HTML]{0000FF} 29.05} & {\color[HTML]{0000FF} 26.81} & 23.75                        & 22.21                        & 21.21                        & 20.51                        \\
                         & EDSR-IPE~\cite{liu2021enhancing}           & {\color[HTML]{0000FF} 34.72} & 31.01                        & 29.04                        & 26.79                        & 23.75                        & 22.21                        & 21.22                        & 20.51                        \\
                         & EDSR-LTE~\cite{lee2022local}           & {\color[HTML]{0000FF} 34.72} & {\color[HTML]{0000FF} 31.02} & 29.04                        & {\color[HTML]{0000FF} 26.81} & {\color[HTML]{0000FF} 23.78} & {\color[HTML]{0000FF} 22.23} & {\color[HTML]{0000FF} 21.24} & {\color[HTML]{0000FF} 20.53} \\
\multirow{-7}{*}{EDSR~\cite{lim2017enhanced}}   & EDSR-LIWT(ours)   & {\color[HTML]{FF0000} 34.79} & {\color[HTML]{FF0000} 31.12} & {\color[HTML]{FF0000} 29.15} & {\color[HTML]{FF0000} 26.91} & {\color[HTML]{FF0000} 23.86} & {\color[HTML]{FF0000} 22.31} & {\color[HTML]{FF0000} 21.30} & {\color[HTML]{FF0000} 20.60} \\ \hline
                         & RDN-baseline~\cite{zhang2018residual}       & 34.94                        & 31.22                        & 29.19                        & -                            & -                            & -                            & -                            & -                            \\
                         & RDN-Meta-SR~\cite{hu2019meta}        & 35.00                        & 31.27                        & 29.25                        & 26.88                        & 23.73                        & 22.18                        & 21.17                        & 20.47                        \\
                         & RDN-LIIF~\cite{chen2021learning}           & 34.99                        & 31.26                        & 29.27                        & 26.99                        & 23.89                        & 22.34                        & 21.31                        & 20.59                        \\
                         & RDN-UltraSR~\cite{xu2021ultrasr}        & 35.00                        & 31.30                        & 29.32                        & 27.03                        & 23.73                        & 22.36                        & 21.33                        & 20.61                        \\
                         & RDN-IPE~\cite{liu2021enhancing}            & {\color[HTML]{0000FF} 35.04} & {\color[HTML]{0000FF} 31.32} & 29.32                        & {\color[HTML]{0000FF} 27.04} & 23.93                        & 22.38                        & 21.34                        & 20.63                        \\
                         & RDN-LTE~\cite{lee2022local}            & {\color[HTML]{0000FF} 35.04} & {\color[HTML]{0000FF} 31.32} & {\color[HTML]{0000FF} 29.33} & {\color[HTML]{0000FF} 27.04} & {\color[HTML]{0000FF} 23.95} & {\color[HTML]{0000FF} 22.40} & {\color[HTML]{0000FF} 21.36} & {\color[HTML]{0000FF} 20.64} \\
\multirow{-7}{*}{RDN~\cite{zhang2018residual}}    & RDN-LIWT(ours)    & {\color[HTML]{FF0000} 35.07} & {\color[HTML]{FF0000} 31.36} & {\color[HTML]{FF0000} 29.39} & {\color[HTML]{FF0000} 27.11} & {\color[HTML]{FF0000} 24.03} & {\color[HTML]{FF0000} 22.47} & {\color[HTML]{FF0000} 21.43} & {\color[HTML]{FF0000} 20.70} \\ \hline
                         & SwinIR-baseline~\cite{liang2021swinir}    & 34.94                        & 31.22                        & 29.19                        & -                            & -                            & -                            & -                            & -                            \\
                         & SwinIR-Meta-SR~\cite{hu2019meta}     & 35.15                        & 31.40                        & 29.33                        & 26.94                        & 23.80                        & 22.26                        & 21.26                        & 20.54                        \\
                         & SwinIR-LIIF~\cite{chen2021learning}        & 35.17                        & 31.46                        & 29.46                        & 27.15                        & 24.02                        & 22.43                        & 21.40                        & 20.67                        \\
                         & SwinIR-LTE~\cite{lee2022local}         & {\color[HTML]{0000FF} 35.24} & {\color[HTML]{0000FF} 31.50} & {\color[HTML]{0000FF} 29.51} & {\color[HTML]{0000FF} 27.20} & {\color[HTML]{0000FF} 24.09} & {\color[HTML]{0000FF} 22.50} & {\color[HTML]{0000FF} 21.47} & {\color[HTML]{0000FF} 20.73} \\
\multirow{-5}{*}{SwinIR~\cite{liang2021swinir}} & SwinIR-LIWT(ours) & {\color[HTML]{FF0000} 35.25} & {\color[HTML]{FF0000} 31.53} & {\color[HTML]{FF0000} 29.55} & {\color[HTML]{FF0000} 27.25} & {\color[HTML]{FF0000} 24.15} & {\color[HTML]{FF0000} 22.56} & {\color[HTML]{FF0000} 21.52} & {\color[HTML]{FF0000} 20.77} \\ \hline
\end{tabular}
}
\end{center}
\caption{Quantitative comparison on the DIV2K validation set~\cite{agustsson2017ntire}, with the best and second best results highlighted in \textcolor{red}{red} and \textcolor{blue}{blue}, respectively.}
\label{table1}
\end{table}
\vspace{-0.9em}
\section{Experimental Results\vspace{-0.2em}}
\subsection{Experimental Setup}
{\bf Dataset.} Following \cite{chen2021learning,lee2022local}, we train our model on the training set of the DIV2K dataset~\cite{agustsson2017ntire}, which comprises 800 images with a resolution of 2K. We evaluate the performance on the DIV2K validation dataset and benchmark datasets, including Set5~\cite{bevilacqua2012low}, Set14~\cite{zeyde2012single}, B100~\cite{martin2001database}, and Urban100~\cite{huang2015single}. Consistent with previous SR methods based on implicit neural representation \cite{chen2021learning,lee2022local,hu2019meta,xu2021ultrasr,liu2021enhancing}, we employed PSNR as the evaluation metric. We also provide SSIM and LPIPS evaluation metrics in the appendix.\\
{\bf Curriculum Learning and Training details.} Consistent with prior work \cite{lee2022local,chen2023cascaded}, we employed EDSR~\cite{lim2017enhanced}, RDN~\cite{zhang2018residual}, and SwinIR~\cite{liang2021swinir} without upsampling modules to serve as encoders. For the synthesis of training data in each batch, we cropped patches of size $48s\times48s$ from each HR image, then downsampled them using bicubic interpolation to produce LR images of size $48\times48$ as inputs to the model. Subsequently, we sampled $48^2$ pixels from the corresponding cropped patches to form RGB-coordinate pairs as ground truth. To enhance the model's ability to recover high-frequency details in large-scale SR, we proposed a curriculum learning training strategy. We trained the model for 1000 epochs. Within the first 250 epochs, we sampled the upsampling factor $s$ from a uniform distribution $U(1,4)$. In the range of epochs from 250 to 500, we sampled the upsampling factor $s$ from a uniform distribution $U(1,6)$. In the range of epochs from 500 to 1000, we sampled the upsampling factor $s$ from a uniform distribution $U(1,8)$. By gradually expanding the range of scale factor sampling during training, the model can progressively learn how to handle large scaling factors. We set the batch size to 32 and utilized the Adam optimizer. The learning rate was initially set to 1e-4 and decayed by a factor of 0.5 every 200 epochs. Please refer to the appendix for our analysis of the training strategy.

\begin{table}[h]
\setlength{\abovecaptionskip}{-0.5cm}
\setlength{\belowcaptionskip}{-0.6cm}
\begin{center}
\resizebox{1.0\columnwidth}{!}{
\begin{tabular}{c|cccccc|cccccc|cccccc|cccccc}
\hline
                          & \multicolumn{6}{c|}{Set5~\cite{bevilacqua2012low}}                                                                                                                                                               & \multicolumn{6}{c|}{Set14~\cite{zeyde2012single}}                                                                                                                                                              & \multicolumn{6}{c|}{B100~\cite{martin2001database}}                                                                                                                                                               & \multicolumn{6}{c}{Urban100~\cite{huang2015single}}                                                                                                                                                            \\ \cline{2-25} 
\multirow{-2}{*}{Methods} & $\times2$                           & $\times3$                           & $\times4$                           & $\times6$                           & $\times8$                           & $\times12$                          & $\times2$                           & $\times3$                           & $\times4$                           & $\times6$                           & $\times8$                           & $\times12$                          & $\times2$                           & $\times3$                           & $\times4$                           & $\times6$                           & $\times8$                           & $\times12$                          & $\times2$                           & $\times3$                           & $\times4$                           & $\times6$                           & $\times8$                           & $\times12$                          \\ \hline
RDN~\cite{zhang2018residual}                       & {\color[HTML]{0000FF} 38.24} & 34.71                        & 32.47                        & -                            & -                            & -                            & 34.01                        & 30.57                        & 28.81                        & -                            & -                            & -                            & 32.34                        & 29.26                        & 27.72                        & -                            & -                            & -                            & 32.89                        & 28.80                        & 26.61                        & -                            & -                            & -                            \\
RDN-Meta-SR~\cite{hu2019meta}               & 38.22                        & 34.63                        & 32.38                        & 29.04                        & 26.96                        & 24.68                        & 33.98                        & 30.54                        & 28.78                        & 26.51                        & 24.97                        & 23.17                        & 32.33                        & 29.26                        & 27.71                        & 25.90                        & 24.83                        & 23.47                        & 32.92                        & 28.82                        & 26.55                        & 23.99                        & 22.59                        & 21.00                        \\
RDN-LIIF~\cite{chen2021learning}                  & 38.17                        & 34.68                        & 32.50                        & 29.15                        & 27.14                        & {\color[HTML]{0000FF} 24.86} & 33.97                        & 30.53                        & 28.80                        & 26.64                        & 25.15                        & 23.24                        & 32.32                        & 29.26                        & 27.74                        & 25.98                        & 24.91                        & 23.57                        & 32.87                        & 28.82                        & 26.68                        & 24.20                        & 22.79                        & 21.15                        \\
RDN-UltraSR~\cite{xu2021ultrasr}               & 38.21                        & 34.67                        & 32.49                        & {\color[HTML]{0000FF} 29.33} & 27.24                        & 24.81                        & 33.97                        & {\color[HTML]{0000FF} 30.59} & 28.86                        & 26.69                        & {\color[HTML]{0000FF} 25.25} & {\color[HTML]{0000FF} 23.32} & 32.35                        & 29.29                        & {\color[HTML]{0000FF} 27.77} & {\color[HTML]{0000FF} 26.01} & {\color[HTML]{0000FF} 24.96} & {\color[HTML]{0000FF} 23.59} & 32.97                        & 28.92                        & 26.78                        & {\color[HTML]{0000FF} 24.30} & 22.87                        & 21.20                        \\
RDN-IPE~\cite{liu2021enhancing}                   & 38.11                        & 34.68                        & 32.51                        & 29.25                        & 27.22                        & -                            & 33.94                        & 30.47                        & 28.75                        & 26.58                        & 25.09                        & -                            & 32.31                        & 29.28                        & 27.76                        & 26.00                        & 24.93                        & -                            & 32.97                        & 28.82                        & 26.76                        & 24.26                        & 22.87                        & -                            \\
RDN-LTE~\cite{lee2022local}                   & 38.23                        & {\color[HTML]{0000FF} 34.72} & {\color[HTML]{0000FF} 32.61} & 29.32                        & {\color[HTML]{0000FF} 27.26} & 24.79                        & {\color[HTML]{0000FF} 34.09} & 30.58                        & {\color[HTML]{0000FF} 28.88} & {\color[HTML]{0000FF} 26.71} & 25.16                        & 23.31                        & {\color[HTML]{0000FF} 32.36} & {\color[HTML]{0000FF} 29.30} & {\color[HTML]{0000FF} 27.77} & {\color[HTML]{0000FF} 26.01} & 24.95                        & 23.60                        & {\color[HTML]{0000FF} 33.04} & {\color[HTML]{0000FF} 28.97} & {\color[HTML]{0000FF} 26.81} & 24.28                        & {\color[HTML]{0000FF} 22.88} & {\color[HTML]{0000FF} 21.22} \\
RDN-LIWT(ours)           & {\color[HTML]{FF0000} 38.28} & {\color[HTML]{FF0000} 34.80} & {\color[HTML]{FF0000} 32.63} & {\color[HTML]{FF0000} 29.45} & {\color[HTML]{FF0000} 27.38} & {\color[HTML]{FF0000} 25.00} & {\color[HTML]{FF0000} 34.20} & {\color[HTML]{FF0000} 30.69} & {\color[HTML]{FF0000} 28.94} & {\color[HTML]{FF0000} 26.80} & {\color[HTML]{FF0000} 25.30} & {\color[HTML]{FF0000} 23.36} & {\color[HTML]{FF0000} 32.39} & {\color[HTML]{FF0000} 29.32} & {\color[HTML]{FF0000} 27.81} & {\color[HTML]{FF0000} 26.05} & {\color[HTML]{FF0000} 24.99} & {\color[HTML]{FF0000} 23.63} & {\color[HTML]{FF0000} 33.11} & {\color[HTML]{FF0000} 29.07} & {\color[HTML]{FF0000} 26.95} & {\color[HTML]{FF0000} 24.41} & {\color[HTML]{FF0000} 23.01} & {\color[HTML]{FF0000} 21.33} \\ \hline
SwinIR~\cite{liang2021swinir}           & {\color[HTML]{0000FF} 38.35} & {\color[HTML]{0000FF} 34.89} & 32.72                        & -                            & -                            & -                            & 34.14                        & 30.77                        & 28.94                        & -                            & -                            & -                            & {\color[HTML]{FF0000} 32.44} & {\color[HTML]{0000FF} 29.37} & 27.83                        & -                            & -                            & -                            & 33.4                         & 29.29                        & 27.07                        & -                            & -                            & -                            \\
SwinIR-Meta-SR~\cite{hu2019meta}            & 38.26                        & 34.77                        & 32.47                        & 29.09                        & 27.02                        & 24.82                        & 34.14                        & 30.66                        & 28.85                        & 26.58                        & 25.09                        & 23.33                        & 32.39                        & 29.31                        & 27.75                        & 25.94                        & 24.86                        & 23.59                        & 33.29                        & 29.12                        & 26.76                        & 24.16                        & 22.75                        & 21.31                        \\
SwinIR-LIIF~\cite{chen2021learning}               & 38.28                        & 34.87                        & 32.73                        & 29.46                        & {\color[HTML]{0000FF} 27.36} & {\color[HTML]{0000FF} 24.99} & 34.14                        & 30.75                        & 28.98                        & 26.82                        & 25.34                        & 23.39                        & 32.39                        & 29.34                        & 27.84                        & 26.07                        & 25.01                        & 23.64                        & 33.36                        & 29.33                        & 27.15                        & 24.59                        & 23.14                        & 21.43                        \\
SwinIR-LTE~\cite{lee2022local}                & 38.33                        & {\color[HTML]{0000FF} 34.89} & {\color[HTML]{0000FF} 32.81} & {\color[HTML]{0000FF} 29.50} & 27.35                        & {\color[HTML]{FF0000} 25.07} & {\color[HTML]{0000FF} 34.25} & {\color[HTML]{0000FF} 30.80} & {\color[HTML]{FF0000} 29.06} & {\color[HTML]{0000FF} 26.86} & {\color[HTML]{0000FF} 25.42} & {\color[HTML]{0000FF} 23.44} & {\color[HTML]{FF0000} 32.44} & {\color[HTML]{FF0000} 29.39} & {\color[HTML]{0000FF} 27.86} & {\color[HTML]{0000FF} 26.09} & {\color[HTML]{0000FF} 25.03} & {\color[HTML]{0000FF} 23.66} & {\color[HTML]{0000FF} 33.50} & {\color[HTML]{0000FF} 29.41} & {\color[HTML]{0000FF} 27.24} & {\color[HTML]{0000FF} 24.62} & {\color[HTML]{0000FF} 23.17} & {\color[HTML]{0000FF} 21.50} \\
SwinIR-LIWT(ours)        & {\color[HTML]{FF0000} 38.37} & {\color[HTML]{FF0000} 34.94} & {\color[HTML]{FF0000} 32.84} & {\color[HTML]{FF0000} 29.60} & {\color[HTML]{FF0000} 27.51} & {\color[HTML]{FF0000} 25.07} & {\color[HTML]{FF0000} 34.31} & {\color[HTML]{FF0000} 30.86} & {\color[HTML]{0000FF} 29.05} & {\color[HTML]{FF0000} 26.90} & {\color[HTML]{FF0000} 25.43} & {\color[HTML]{FF0000} 23.48} & {\color[HTML]{0000FF} 32.43} & {\color[HTML]{FF0000} 29.39} & {\color[HTML]{FF0000} 27.88} & {\color[HTML]{FF0000} 26.13} & {\color[HTML]{FF0000} 25.06} & {\color[HTML]{FF0000} 23.69} & {\color[HTML]{FF0000} 33.52} & {\color[HTML]{FF0000} 29.46} & {\color[HTML]{FF0000} 27.30} & {\color[HTML]{FF0000} 24.71} & {\color[HTML]{FF0000} 23.24} & {\color[HTML]{FF0000} 21.57} \\ \hline
\end{tabular}
}
\end{center}
\caption{Quantitative comparison on the benchmark datasets, with the best and second best results highlighted in \textcolor{red}{red} and \textcolor{blue}{blue}, respectively.}
\label{table2}
\end{table}
\vspace{-0.0em}
\subsection{Comparison with Previous Methods\vspace{-0.0em}}
We compare the proposed LIWT with existing arbitrary-scale SR methods \cite{chen2021learning,lee2022local,hu2019meta,xu2021ultrasr,liu2021enhancing}. Table \ref{table1} presents the PSNR metric results of all comparing methods on the DIV2K validation dataset. As shown in Table \ref{table1}, our LIWT achieves the best performance across different scaling factors, which indicates that our method can be integrated with different encoders to improve performance. Table \ref{table2} depicts the results of LIWT on the four benchmark datasets. Our LIWT achieves the best performance across scaling factors of $\times6$, $\times8$, and $\times12$. At low scaling factors of $\times2$, $\times3$, and $\times4$, our method also achieves the performance in most scenarios. We have also provided a quantitative comparison under non-integer scaling factors in the appendix. Figure \ref{fig4} shows the visual results of different methods. Benefiting from the high-frequency prior introduced by wavelet transform in the attention mechanism and the curriculum learning strategy, our method achieves accurate structural restoration.

\begin{figure}[h]
\setlength{\abovecaptionskip}{-0.5cm}
\setlength{\belowcaptionskip}{-0.3cm}
\centerline{\includegraphics[width=\columnwidth]{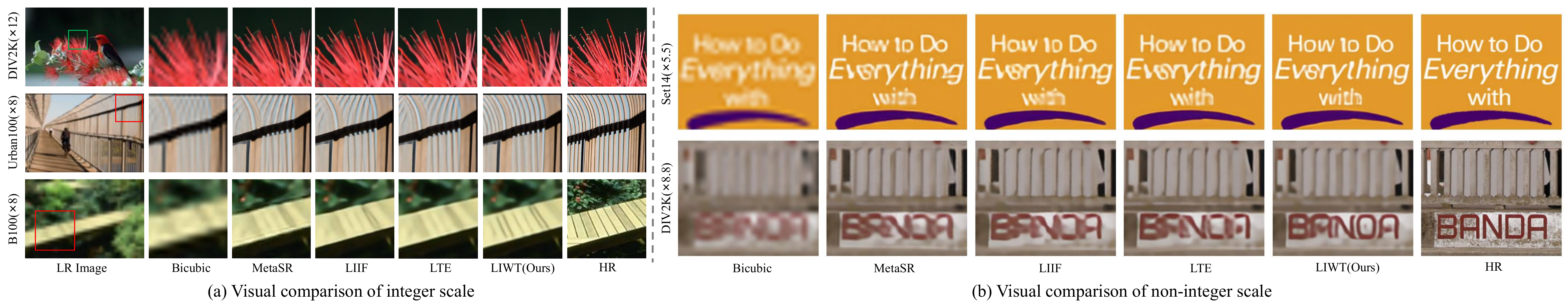}}
\caption{Visual comparison of MetaSR~\cite{hu2019meta}, LIIF~\cite{chen2021learning}, LTE~\cite{lee2022local}, and LIWT using RDN~\cite{zhang2018residual} as the encoder. Zoom in for best view.}
\label{fig4}
\end{figure}


\begin{table}[h]
\setlength{\abovecaptionskip}{-0.3cm}
\setlength{\belowcaptionskip}{-0.5cm}
\begin{center}
\resizebox{0.8\columnwidth}{!}{
\begin{tabular}{|cccc|ccccc|ccccc|}
\hline
\multicolumn{4}{|c|}{Model} & \multicolumn{5}{c|}{DIV2K}                                                                                                                                                                            & \multicolumn{5}{c|}{B100}                                                                                                                                                                             \\ \hline
$F^L$  & $F^H$  & $WHFERB$  & $WERB$  & $\times2$                                    & $\times3$                                    & $\times4$                                    & $\times6$                                    & $\times8$                                    & $\times2$                                    & $\times3$                                    & $\times4$                                    & $\times6$                                    & $\times8$                                    \\ \hline
$\checkmark$   & $\checkmark$   &         &       & 34.76                                 & 31.09                                 & 29.13                                 & 26.89                                 & 25.53                                 & 32.20                                 & 29.15                                 & 27.65                                 & 25.92                                 & 24.87                                 \\
    & $\checkmark$   &         & $\checkmark$     & 34.77                                 & 31.11                                 & 29.14                                 & 26.90                                 & 25.54                                 & 32.22                                 & 29.16                                 & 27.65                                 & 25.92                                 & 24.87                                 \\
$\checkmark$   &     & $\checkmark$       & $\checkmark$     & 34.78                                 & 31.11                                 & 29.14                                 & {\color[HTML]{000000} \textbf{26.91}} & 25.54                                 & 32.22                                 & {\color[HTML]{000000} \textbf{29.17}} & {\color[HTML]{000000} \textbf{27.67}} & {\color[HTML]{000000} \textbf{25.93}} & {\color[HTML]{000000} \textbf{24.88}} \\
$\checkmark$   & $\checkmark$   &         & $\checkmark$     & {\color[HTML]{000000} \textbf{34.79}} & {\color[HTML]{000000} \textbf{31.12}} & {\color[HTML]{000000} \textbf{29.15}} & {\color[HTML]{000000} \textbf{26.91}} & 25.54                                 & 32.22                                 & 29.16                                 & 27.66                                 & {\color[HTML]{000000} \textbf{25.93}} & 24.87                                 \\
$\checkmark$   & $\checkmark$   & $\checkmark$       & $\checkmark$     & {\color[HTML]{000000} \textbf{34.79}} & {\color[HTML]{000000} \textbf{31.12}} & {\color[HTML]{000000} \textbf{29.15}} & {\color[HTML]{000000} \textbf{26.91}} & {\color[HTML]{000000} \textbf{25.55}} & {\color[HTML]{000000} \textbf{32.23}} & {\color[HTML]{000000} \textbf{29.17}} & {\color[HTML]{000000} \textbf{27.67}} & {\color[HTML]{000000} \textbf{25.93}} & {\color[HTML]{000000} \textbf{24.88}} \\ \hline
\end{tabular}
}
\end{center}
\caption{Ablation study of WERM on the DIV2K validation datasets~\cite{agustsson2017ntire} and B100~\cite{martin2001database}. The best performing results are indicated in \color[HTML]{000000} \textbf{bold}.}
\label{table3}
\end{table}

\subsection{Ablation Study}
\begin{figure}[t]
\setlength{\abovecaptionskip}{-0.5cm}
\setlength{\belowcaptionskip}{-0.1cm}
\centerline{\includegraphics[width=\columnwidth]{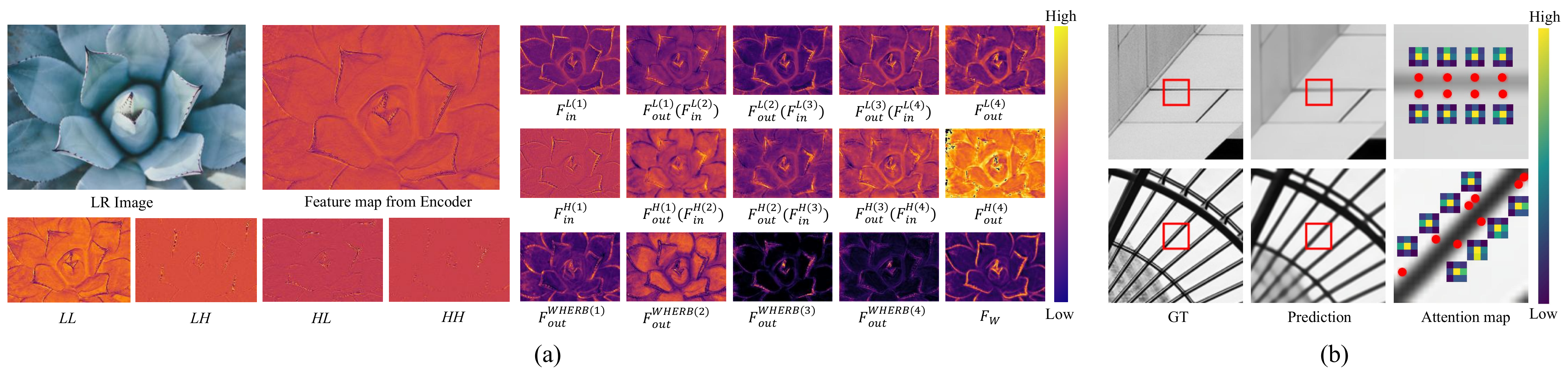}}
\caption{(a) Visualization of feature maps for each stage of WERM; (b) Attention map visualization for WIA.}
\label{fig5}
\end{figure}

{\bf Effectiveness of WERM.} We conduct extensive experiments using EDSR as the encoder to evaluate the effectiveness of WERM, as shown in Table \ref{table3}. Upon separately removing the branches for processing $F^H$ and $F^L$, we observe a decrease in performance, indicating the impact of different frequency bands on reconstruction. Similarly, performance degradation is observed upon removing WERM and WHFERB. Figure \ref{fig5}(a) visualizes the features at various stages of WERM. The output $F_W$ of WERM significantly enhances the representation of sharp edge textures (petal edges), with clear distinctions between edge and smooth regions, achieving more prominent discriminability than any feature at previous stages.

\begin{table}[t!]
\setlength{\abovecaptionskip}{-0.3cm}
\setlength{\belowcaptionskip}{-0.3cm}
\begin{center}
\resizebox{0.8\columnwidth}{!}{
\begin{tabular}{|c|ccccc|ccccc|}
\hline
                          & \multicolumn{5}{c|}{Set14}                                                                                                                                                                            & \multicolumn{5}{c|}{Urban100}                                                                                                                                                                         \\ \cline{2-11} 
\multirow{-2}{*}{Methods} & $\times2$                                    & $\times3$                                    & $\times4$                                    & $\times6$                                    & $\times8$                                    & $\times2$                                    & $\times3$                                    & $\times4$                                    & $\times6$                                    & $\times8$                                    \\ \hline
LIWT                      & 33.73                                 & {\color[HTML]{000000} \textbf{30.48}} & {\color[HTML]{000000} \textbf{28.75}} & {\color[HTML]{000000} \textbf{26.59}} & {\color[HTML]{000000} \textbf{25.10}} & 32.44                                 & {\color[HTML]{000000} \textbf{28.51}} & {\color[HTML]{000000} \textbf{26.44}} & {\color[HTML]{000000} \textbf{24.08}} & {\color[HTML]{000000} \textbf{22.71}} \\ \hline
LIWT(w/o WMPF)            & {\color[HTML]{000000} \textbf{33.77}} & 30.45                                 & 28.73                                 & 26.58                                 & 25.08                                 & {\color[HTML]{000000} \textbf{32.45}} & {\color[HTML]{000000} \textbf{28.51}} & {\color[HTML]{000000} \textbf{26.44}} & 24.03                                 & 22.68                                 \\ \hline
\end{tabular}
}
\end{center}
\caption{Ablation study of WMPF on Set14~\cite{zeyde2012single} and Urban100~\cite{huang2015single}. The best performing results are indicated in \color[HTML]{000000} \textbf{bold}.}
\label{table4}
\end{table}

\begin{table}[t!]
\setlength{\abovecaptionskip}{-0.4cm}
\setlength{\belowcaptionskip}{-0.2cm}
\begin{center}
\resizebox{0.7\columnwidth}{!}{
\begin{tabular}{|c|ccccc|ccccc|}
\hline
                          & \multicolumn{5}{c|}{DIV2K}                                                                                                                                                                            & \multicolumn{5}{c|}{Urban100}                                                                                                                                                                         \\ \cline{2-11} 
\multirow{-2}{*}{Methods} & $\times2$                                    & $\times3$                                    & $\times4$                                    & $\times6$                                    & $\times8$                                    & $\times2$                                    & $\times3$                                    & $\times4$                                    & $\times6$                                    & $\times8$                                    \\ \hline
LIWT                      & {\color[HTML]{000000} \textbf{34.79}} & {\color[HTML]{000000} \textbf{31.12}} & {\color[HTML]{000000} \textbf{29.15}} & {\color[HTML]{000000} \textbf{26.91}} & {\color[HTML]{000000} \textbf{25.55}} & {\color[HTML]{000000} \textbf{32.44}} & {\color[HTML]{000000} \textbf{28.51}} & 26.44                                 & {\color[HTML]{000000} \textbf{24.08}} & {\color[HTML]{000000} \textbf{22.71}} \\ \hline
LIWT(SA)                  & {\color[HTML]{000000} \textbf{34.79}} & 31.11                                 & 29.14                                 & {\color[HTML]{000000} \textbf{26.91}} & {\color[HTML]{000000} \textbf{25.55}} & 32.43                                 & 28.49                                 & {\color[HTML]{000000} \textbf{26.45}} & 24.03                                 & 22.68                                 \\
LIWT(WSA)                 & {\color[HTML]{000000} \textbf{34.79}} & 31.11                                 & {\color[HTML]{000000} \textbf{29.15}} & {\color[HTML]{000000} \textbf{26.91}} & 25.54                                 & 32.42                                 & 28.48                                 & {\color[HTML]{000000} \textbf{26.45}} & 24.05                                 & 22.70                                 \\ \hline
\end{tabular}
}
\end{center}
\caption{Ablation study of WIA on the DIV2K validation set~\cite{agustsson2017ntire} and Urban100~\cite{huang2015single}. The best performing results are indicated in \color[HTML]{000000} \textbf{bold}.}
\label{table5}
\end{table}

\noindent {\bf Effectiveness of WMPF and WIA.} We conduct extensive experiments on different datasets to assess the effectiveness of WMPF and WIA. Table \ref{table4} demonstrates a performance decrease at most scaling factors after removing WMPF, indicating the role of high-frequency priors introduced by WMPF. To investigate the effectiveness of the cross-attention mechanism in WIA, we design the following two variant networks based on the self-attention mechanism: (1) LIWT(SA): replacing the query $q$ with the latent embedding obtained by projecting the features Z from the encoder; (2) LIWT(WSA): replacing both the query $q$ and key $k$ with the latent embedding obtained by projecting $F_{W}^{\uparrow}$. Table \ref{table5} shows that LIWT achieves the best results across various scaling factors except for $\times4$ on Urban100 only. We provide visualizations of the cross-attention maps in Figure \ref{fig5}(b) to visually illustrate the effectiveness of WIA. The direction of attention map distribution aligns with the direction of these high-frequency details. This indicates that WIA successfully captures sharp-edge detail features. Besides, we observe a performance drop by removing the positional encoding and cell in Table \ref{table6}, highlighting the importance of position and pixel size information for LIWT.

\begin{table}[t!]
\setlength{\abovecaptionskip}{-0.3cm}
\setlength{\belowcaptionskip}{-0.3cm}
\begin{center}
\resizebox{0.7\columnwidth}{!}{
\begin{tabular}{|c|ccccccccc|}
\hline
Methods        & $\times2$                                    & $\times3$                                    & $\times4$                                    & $\times6$                                    & $\times8$                                    & $\times12$                                   & $\times24$                                   & $\times27$                                   & $\times30$                                  \\ \hline
LIWT           & {\color[HTML]{000000} \textbf{34.79}} & {\color[HTML]{000000} \textbf{31.12}} & {\color[HTML]{000000} \textbf{29.15}} & {\color[HTML]{000000} \textbf{26.91}} & {\color[HTML]{000000} \textbf{25.55}} & {\color[HTML]{000000} \textbf{23.86}} & {\color[HTML]{000000} \textbf{21.30}} & {\color[HTML]{000000} \textbf{20.92}} & {\color[HTML]{000000} \textbf{20.6}} \\ \hline
LIWT(w/o PE)   & 31.35                                 & 28.32                                 & 26.69                                 & 24.83                                 & 23.69                                 & 22.27                                 & 20.20                                 & 19.86                                 & 19.60                                \\
LIWT(w/o Cell) & 34.67                                 & 31.06                                 & 29.13                                 & 26.90                                 & {\color[HTML]{000000} \textbf{25.55}} & {\color[HTML]{000000} \textbf{23.86}} & 21.29                                 & 20.90                                 & 20.58                                \\ \hline
\end{tabular}
}
\end{center}
\caption{Ablation study of positional encoding in WIA, and cell in LIWT on the DIV2K validation set~\cite{agustsson2017ntire}. The best performing results are indicated in \color[HTML]{000000} \textbf{bold}.} 
\label{table6}
\end{table}

\noindent {\bf Complexity analysis.} We investigate the computational consumption on an NVIDIA RTX 3090 24GB device. We use LR images of $192\times192$ size as input and evaluated them in the $\times4$ SR task. As shown in Table \ref{table7}, although our model is slightly inferior to LTE~\cite{lee2022local} and LIIF~\cite{chen2021learning} in terms of model size, memory consumption, and inference time, it can bring a PSNR improvement of more than 0.1 dB.

\begin{table}[t!]
\setlength{\abovecaptionskip}{-0.2cm}
\setlength{\belowcaptionskip}{-0.3cm}
\begin{center}
\resizebox{0.6\columnwidth}{!}{
\begin{tabular}{c|ccccc}
\hline
Method & Eval/Query & Params & Mem(GB) & Time(s) & PSNR  \\ \hline
LIIF~\cite{chen2021learning}   & 96×96      & 1.57M  & 2.5     & 0.28    & 27.53 \\
LTE~\cite{lee2022local}    & 96×96      & 1.71M  & 2.5     & 0.26    & 27.57 \\ \hline
Ours   & 96×96      & 3.97M  & 3.1     & 0.50    & 27.68 \\ \hline
\end{tabular}
}
\end{center}
\caption{Comparisons of computational consumption of different methods.}
\label{table7}
\end{table}
\vspace{-1.0em}
\section{Conclusion\vspace{-0.3em}}
In this paper, we propose a novel Local Implicit Wavelet Transformer for arbitrary-scale SR. Specifically, we introduce high-frequency prior information of the features via discrete wavelet transform and a Wavelet Enhancement Residual Module and then effectively utilize high-frequency priori information to improve the reconstruction performance by utilizing the proposed Wavelet Mutual Projected Fusion and Wavelet-aware Implicit Attention module. Extensive experiments have shown that LIWT has superior performance to other methods.

\appendix
\section{Appendix}
\subsection{Haar Wavelet Transform}
\label{sec:1}
The Haar wavelet transform is a simple and computationally efficient method for decomposing input signals into low-frequency and high-frequency sub-bands, widely employed in the field of computer vision ~\cite{guo2017deep,hsu2023wavelet,liu2019multi,xin2020wavelet,xue2020wavelet,zou2022joint}. In this paper, we utilize the Haar wavelet transform to perform the discrete wavelet transform (DWT) on the features $Z$ obtained from the encoder. The Haar wavelet transform typically involves processing the input signal with high-pass filter $H^T$ and low-pass filter $L^T$ to obtain different sub-bands. Specifically, the low-pass and high-pass filters are:
\begin{equation}
\mathbf{L}^{\mathbf{T}}=\frac{1}{\sqrt{2}}\left[\begin{array}{ll}
1 & 1
\end{array}\right], \quad \mathbf{H}^{\mathbf{T}}=\frac{1}{\sqrt{2}}\left[\begin{array}{ll}
-1 & 1
\end{array}\right] 
\end{equation}
Similarly, the filters of the Haar wavelet transform consist of four 2×2 kernels, including $LL^{T}$, $LH^{T}$, $HL^{T}$ and $HH^{T}$. In this paper, we use $LL^{T}$ to process the feature $Z$ to obtain the low-frequency component $LL$, and respectively use $LH^{T}$, $HL^{T}$ and $HH^{T}$ to process the feature $Z$ to obtain the high-frequency components $LH$, $HL$ and $HH$. Following the wavelet transform, the low-frequency component exhibits smooth surface and texture information, while the high-frequency components capture more complex texture details. We denote the low-frequency component $LL$ as $F^L$ and concatenate the high-frequency components $LH$, $HL$, and $HH$ along the channel axis, represented as $F^H$.

\subsection{Analysis on the training strategy}
\label{sec:2}
We analyzed the impact of training strategies on the DIV2K dataset~\cite{agustsson2017ntire}. As shown in Table \ref{table8}, training within a small scale sampling range of $s \sim U(1,4)$ enables the model to achieve good performance at small upscaling scales but sacrifices reconstruction accuracy at higher scaling factors. Expanding the scale sampling range to $s \sim U(1,8)$ during training can enhance the model's reconstruction performance at larger upscaling scales but decrease performance at smaller upscaling scales. Our proposed curriculum learning strategy gradually expands the sampling range during training. Although the performance at scaling factors of $\times2$ and $\times3$ is not as good as training within the small-scale sampling range of $s \sim U(1,4)$, overall, it achieves the best balance across different scaling factors. Our training approach ensures effective reconstruction at large sampling scales and achieves the most optimal or suboptimal results across various scaling factors. To validate the generality of the proposed curriculum learning training strategy, we applied the same training setup to LIIF~\cite{chen2021learning} and LTE~\cite{lee2022local}. As shown in Table \ref{table9}, we observed performance improvements, indicating the effectiveness and generalizability of our training strategy.

\begin{table}[h]
\begin{center}
\resizebox{0.8\columnwidth}{!}{
\begin{tabular}{c|ccccccccc}
\hline
Training strategy            & $\times2$                           & $\times3$                           & $\times4$                           & $\times6$                           & $\times8$                           & $\times12$                          & $\times24$                          & $\times27$                          & $\times30$                          \\ \hline
Curriculum learning strategy & {\color[HTML]{0000FF} 34.79} & {\color[HTML]{0000FF} 31.12} & {\color[HTML]{FF0000} 29.15} & {\color[HTML]{FF0000} 26.91} & {\color[HTML]{FF0000} 25.55} & {\color[HTML]{FF0000} 23.86} & {\color[HTML]{0000FF} 21.3}  & {\color[HTML]{FF0000} 20.92} & {\color[HTML]{FF0000} 20.6}  \\ \hline
Training with $U(1,4)$               & {\color[HTML]{FF0000} 34.84} & {\color[HTML]{FF0000} 31.13} & {\color[HTML]{0000FF} 29.14} & {\color[HTML]{0000FF} 26.89} & {\color[HTML]{0000FF} 25.52} & {\color[HTML]{0000FF} 23.83} & 21.27                        & {\color[HTML]{0000FF} 20.89} & {\color[HTML]{0000FF} 20.57} \\
Training with $U(1,8)$                & 34.78                        & 31.11                        & {\color[HTML]{0000FF} 29.14} & {\color[HTML]{FF0000} 26.91} & {\color[HTML]{FF0000} 25.55} & {\color[HTML]{FF0000} 23.86} & {\color[HTML]{FF0000} 21.31} & {\color[HTML]{FF0000} 20.92} & {\color[HTML]{FF0000} 20.6}  \\ \hline
\end{tabular}
}
\end{center}
\caption{The average PSNR (dB) of different training strategies on the DIV2K validation set~\cite{agustsson2017ntire}.}
\label{table8}
\end{table}

\begin{table}[h]
\begin{center}
\resizebox{0.9\columnwidth}{!}{
\begin{tabular}{c|c|cccccc}
\hline
Training strategy           & Method    & $\times2$           & $\times3$           & $\times4$           & $\times6$           & $\times12$          & $\times18$          \\ \hline
\multirow{2}{*}{Original}   & EDSR-LIIF~\cite{chen2021learning} & 34.67        & 30.96        & 29           & 26.75        & 23.71        & 22.17        \\
                            & EDSR-LTE~\cite{lee2022local}  & 34.72        & 31.02        & 29.04        & 26.81        & 23.78        & 22.23        \\ \hline
\multirow{2}{*}{Curriculum} & EDSR-LIIF~\cite{chen2021learning} & 34.65(\color[HTML]{000000} \textbf{-0.02}) & 30.99(\color[HTML]{000000} \textbf{+0.03}) & 29.05(\color[HTML]{000000} \textbf{+0.05}) & 26.81(\color[HTML]{000000} \textbf{+0.06}) & 23.77(\color[HTML]{000000} \textbf{+0.06}) & 22.22(\color[HTML]{000000} \textbf{+0.05}) \\
                            & EDSR-LTE~\cite{lee2022local}  & 34.7(\color[HTML]{000000} \textbf{-0.02})  & 31.03(\color[HTML]{000000} \textbf{+0.01}) & 29.07(\color[HTML]{000000} \textbf{+0.03}) & 26.85(\color[HTML]{000000} \textbf{+0.04}) & 23.82(\color[HTML]{000000} \textbf{+0.04}) & 22.28(\color[HTML]{000000} \textbf{+0.05}) \\ \hline
\end{tabular}
}
\end{center}
\caption{The average PSNR (dB) of LIIF~\cite{chen2021learning} and LTE~\cite{lee2022local} training with and without curriculum learning on the DIV2K validation set~\cite{agustsson2017ntire}.}
\label{table9}
\end{table}

\vspace{-1.6em}
\subsection{More evaluation metrics}
We employ two metrics, SSIM and LPIPS, to further demonstrate the effectiveness of LIWT compared to other arbitrary-scale SR methods. We compare the performance of LIWT, LTE~\cite{lee2022local}, LIIF~\cite{chen2021learning}, and MetaSR~\cite{hu2019meta} using SwinIR as the encoder on Set14~\cite{zeyde2012single} and Urban100~\cite{huang2015single} datasets. Typically, higher SSIM and lower LPIPS correspond to better performance. From the results in Table \ref{table10}, it can be observed that except for $\times2$ scaling, our method achieves the highest SSIM and the lowest LPIPS. This indicates that our approach can recover more structural information and has better perceptual quality.

\begin{table}[h]
\begin{center}
\resizebox{0.9\columnwidth}{!}{
\begin{tabular}{|c|c|cc|cc|cc|cc|cc|}
\hline
\multirow{2}{*}{Dataset}  & \multirow{2}{*}{Method} & \multicolumn{2}{c|}{×2}         & \multicolumn{2}{c|}{×3}         & \multicolumn{2}{c|}{×4}         & \multicolumn{2}{c|}{×6}         & \multicolumn{2}{c|}{×8}         \\ \cline{3-12} 
                          &                         & SSIM↑          & LPIPS↓         & SSIM↑          & LPIPS↓         & SSIM↑          & LPIPS↓         & SSIM↑          & LPIPS↓         & SSIM↑          & LPIPS↓         \\ \hline
\multirow{4}{*}{Set14~\cite{zeyde2012single}}    & Meta-SR~\cite{hu2019meta}                 & 0.923          & 0.134          & 0.850          & 0.227          & 0.791          & 0.291          & 0.704          & 0.382          & 0.648          & 0.446          \\
                          & LIIF~\cite{chen2021learning}                    & 0.923          & 0.134          & 0.851          & 0.227          & 0.792          & 0.293          & 0.707          & 0.380          & 0.652          & 0.443          \\
                          & LTE~\cite{lee2022local}                     & \textbf{0.924} & 0.133          & 0.852          & 0.224          & 0.794          & 0.292          & 0.709          & 0.377          & 0.655          & 0.440          \\
                          & LIWT(Ours)                    & \textbf{0.924} & \textbf{0.132} & \textbf{0.853} & \textbf{0.223} & \textbf{0.795} & \textbf{0.289} & \textbf{0.712} & \textbf{0.373} & \textbf{0.657} & \textbf{0.436} \\ \hline
\multirow{4}{*}{Urban100~\cite{huang2015single}} & Meta-SR~\cite{hu2019meta}                 & 0.939          & 0.102          & 0.873          & 0.186          & 0.810          & 0.251          & 0.709          & 0.347          & 0.638          & 0.415          \\
                          & LIIF~\cite{chen2021learning}                    & 0.939          & 0.102          & 0.876          & 0.188          & 0.817          & 0.258          & 0.719          & 0.349          & 0.650          & 0.415          \\
                          & LTE~\cite{lee2022local}                     & \textbf{0.941} & 0.100          & 0.877          & 0.186          & 0.820          & 0.254          & 0.722          & 0.343          & 0.653          & 0.408          \\
                          & LIWT(Ours)                    & \textbf{0.941} & \textbf{0.099} & \textbf{0.878} & \textbf{0.183} & \textbf{0.821} & \textbf{0.250} & \textbf{0.726} & \textbf{0.336} & \textbf{0.657} & \textbf{0.401} \\ \hline
\end{tabular}
}
\end{center}
\caption{Comparison of more evaluation metrics (SSIM$\uparrow$ and LPIPS$\downarrow$) on Set14~\cite{zeyde2012single} and Urban100~\cite{huang2015single}.}
\label{table10}
\end{table}

\vspace{-1.6em}
\subsection{Comparison with DWT-based SR methods}
We compare LIWT with other DWT-based SR methods~\cite{guo2017deep,jeevan2024wavemixsr,liu2019multi,xin2020wavelet,xue2020wavelet,zou2022joint} using PSNR and SSIM metrics on Set14~\cite{zeyde2012single} and Urban100~\cite{huang2015single}, where LIWT utilizes SwinIR\cite{liang2021swinir} as the encoder. As shown in Table \ref{table11}, our LIWT achieves the best results at various scaling factors.

\begin{table}[h]
\begin{center}
\resizebox{0.6\columnwidth}{!}{
\begin{tabular}{|c|c|cc|cc|cc|}
\hline
                           &                          & \multicolumn{2}{c|}{×2}                                     & \multicolumn{2}{c|}{×3}                                     & \multicolumn{2}{c|}{×4}                                     \\ \cline{3-8} 
\multirow{-2}{*}{Dataset}  & \multirow{-2}{*}{Method} & \multicolumn{1}{c|}{PSNR↑}   & SSIM↑                        & \multicolumn{1}{c|}{PSNR↑}   & SSIM↑                        & \multicolumn{1}{c|}{PSNR↑}   & SSIM↑                        \\ \hline
                           & MWCNN~\cite{liu2019multi}                    & 33.71                        & 0.918                        & 30.14                        & 0.841                        & 28.58                        & 0.788                        \\
                           & DWSR~\cite{guo2017deep}                     & 33.07                        & 0.911                        & 29.83                        & 0.831                        & 28.04                        & 0.767                        \\
                           & WRAN~\cite{xue2020wavelet}                     & {\color[HTML]{0000FF} 34.21} & 0.922                        & {\color[HTML]{0000FF} 30.71} & {\color[HTML]{0000FF} 0.852} & 28.60                        & 0.786                        \\
                           & WDRN~\cite{xin2020wavelet}                     & 33.90                        & 0.921                        & 30.50                        & 0.845                        & 28.75                        & 0.786                        \\
                           & JWSGN~\cite{zou2022joint}                    & 34.17                        & {\color[HTML]{0000FF} 0.923} & -                            & -                            & {\color[HTML]{0000FF} 28.96} & {\color[HTML]{0000FF} 0.789} \\
                           & WaveMixSR~\cite{jeevan2024wavemixsr}                & 31.27                        & 0.904                        & 28.77                        & 0.841                        & 26.25                        & 0.751                        \\
\multirow{-7}{*}{Set14~\cite{zeyde2012single}}    & LIWT(Ours)                     & {\color[HTML]{FF0000} 34.31} & {\color[HTML]{FF0000} 0.924} & {\color[HTML]{FF0000} 30.86} & {\color[HTML]{FF0000} 0.853} & {\color[HTML]{FF0000} 29.05} & {\color[HTML]{FF0000} 0.795} \\ \hline
                           & MWCNN~\cite{liu2019multi}                    & 32.36                        & 0.931                        & 28.19                        & 0.852                        & 26.37                        & 0.789                        \\
                           & DWSR~\cite{guo2017deep}                     & 30.46                        & 0.916                        & -                            & -                            & 25.26                        & 0.755                        \\
                           & WRAN~\cite{xue2020wavelet}                     & {\color[HTML]{0000FF} 33.47} & {\color[HTML]{0000FF} 0.940} & {\color[HTML]{0000FF} 28.99} & {\color[HTML]{0000FF} 0.869} & 26.74                        & 0.803                        \\
                           & WDRN~\cite{xin2020wavelet}                     & 32.64                        & 0.937                        & 28.59                        & 0.862                        & 26.41                        & 0.797                        \\
                           & JWSGN~\cite{zou2022joint}                    & 33.17                        & 0.938                        & -                            & -                            & {\color[HTML]{0000FF} 26.82} & {\color[HTML]{0000FF} 0.807} \\
                           & WaveMixSR~\cite{jeevan2024wavemixsr}                & 29.14                        & 0.908                        & 25.82                        & 0.819                        & 23.57                        & 0.730                        \\
\multirow{-7}{*}{Urban100~\cite{huang2015single}} & LIWT(Ours)                     & {\color[HTML]{FF0000} 33.52} & {\color[HTML]{FF0000} 0.941} & {\color[HTML]{FF0000} 29.46} & {\color[HTML]{FF0000} 0.878} & {\color[HTML]{FF0000} 27.30} & {\color[HTML]{FF0000} 0.821} \\ \hline
\end{tabular}
}
\end{center}
\caption{Comparison of PSNR$\uparrow$ and SSIM$\uparrow$ for different DWT-based methods on Set14~\cite{zeyde2012single} and Urban100~\cite{huang2015single}.}
\label{table11}
\end{table}

\subsection{Comparison of different arbitrary-scale SR methods at non-integer scale}
To further assess the advantages of our method over other arbitrary-scale SR methods, we present comparative results of PSNR and SSIM metrics at non-integer scaling factors on Set14~\cite{zeyde2012single} and Urban100~\cite{huang2015single}. As shown in Table \ref{table12}, Our LIWT achieves optimal results at various scaling factors.

\begin{table}[h]
\begin{center}
\resizebox{0.9\columnwidth}{!}{
\begin{tabular}{|c|c|cc|cc|cc|cc|cc|cc|cc|cc|}
\hline
\multirow{2}{*}{Dataset}  & \multirow{2}{*}{Method} & \multicolumn{2}{c|}{$\times2.2$}       & \multicolumn{2}{c|}{$\times2.5$}       & \multicolumn{2}{c|}{$\times3.3$}       & \multicolumn{2}{c|}{$\times3.5$}       & \multicolumn{2}{c|}{$\times4.4$}       & \multicolumn{2}{c|}{$\times5.5$}       & \multicolumn{2}{c|}{$\times6.6$}       & \multicolumn{2}{c|}{$\times7.7$}       \\ \cline{3-18} 
                          &                         & PSNR↑          & SSIM↑          & PSNR↑          & SSIM↑          & PSNR↑          & SSIM↑          & PSNR↑          & SSIM↑          & PSNR↑          & SSIM↑          & PSNR↑          & SSIM↑          & PSNR↑          & SSIM↑          & PSNR↑          & SSIM↑          \\ \hline
\multirow{4}{*}{Set14~\cite{zeyde2012single}}    & Meta-SR~\cite{hu2019meta}                 & 31.84          & 0.899          & 31.48          & 0.883          & 29.14          & 0.825          & 29.26          & 0.813          & 28.18          & 0.768          & 26.92          & 0.720          & 26.14          & 0.684          & 25.32          & 0.654          \\
                          & LIIF~\cite{chen2021learning}                    & 31.91          & 0.899          & 31.54          & 0.884          & 29.44          & 0.825          & 29.29          & 0.814          & 28.25          & 0.770          & 27.02          & 0.722          & 26.25          & 0.688          & 25.41          & 0.658          \\
                          & LTE~\cite{lee2022local}                     & 31.93          & 0.900          & 31.55          & 0.884          & 29.48          & 0.826          & 29.30          & 0.815          & 28.29          & 0.771          & 27.07          & 0.724          & 26.30          & 0.689          & \textbf{25.50} & 0.661          \\
                          & LIWT(Ours)                    & \textbf{31.95} & \textbf{0.900} & \textbf{31.60} & \textbf{0.885} & \textbf{29.49} & \textbf{0.827} & \textbf{29.34} & \textbf{0.816} & \textbf{28.31} & \textbf{0.772} & \textbf{27.08} & \textbf{0.725} & \textbf{26.33} & \textbf{0.691} & \textbf{25.50} & \textbf{0.662} \\ \hline
\multirow{4}{*}{Urban100~\cite{huang2015single}} & Meta-SR~\cite{hu2019meta}                 & 31.72          & 0.923          & 28.37          & 0.884          & 28.07          & 0.851          & 27.45          & 0.836          & 25.96          & 0.785          & 24.71          & 0.730          & 23.76          & 0.684          & 23.01          & 0.645          \\
                          & LIIF~\cite{chen2021learning}                    & 31.78          & 0.923          & 28.48          & 0.885          & 28.26          & 0.854          & 27.67          & 0.840          & 26.24          & 0.792          & 24.96          & 0.739          & 23.97          & 0.694          & 23.18          & 0.656          \\
                          & LTE~\cite{lee2022local}                     & 31.92          & 0.925          & \textbf{28.51} & 0.887          & 28.35          & 0.856          & 27.76          & 0.842          & 26.33          & 0.796          & 25.04          & 0.742          & 24.04          & 0.698          & 23.21          & 0.659          \\
                          & LIWT(Ours)                    & \textbf{31.95} & \textbf{0.925} & \textbf{28.51} & \textbf{0.886} & \textbf{28.41} & \textbf{0.857} & \textbf{27.81} & \textbf{0.843} & \textbf{26.39} & \textbf{0.797} & \textbf{25.13} & \textbf{0.746} & \textbf{24.12} & \textbf{0.701} & \textbf{23.29} & \textbf{0.662} \\ \hline
\end{tabular}
}
\end{center}
\caption{Comparison of different arbitrary-scale SR methods at non-integer scale on Set14~\cite{zeyde2012single} and Urban100~\cite{huang2015single} (PSNR (dB)).}
\label{table12}
\end{table}

\vspace{-1.6em}
\subsection{The analysis of large-scale SR}
\begin{figure}[h]
\centerline{\includegraphics[width=\columnwidth]{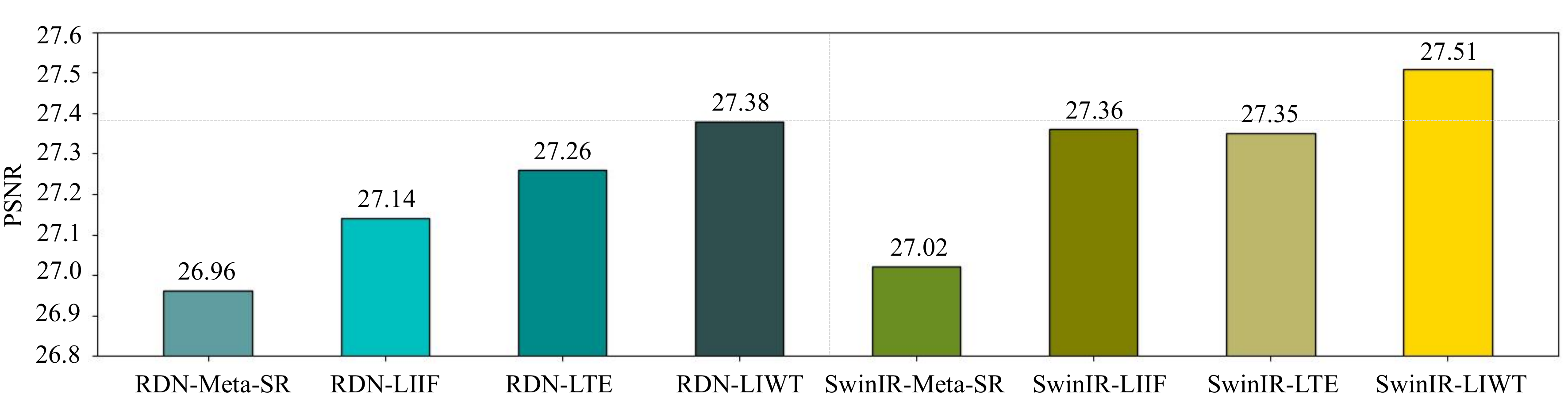}}
\caption{Comparison of $\times8$ SR task on Set5~\cite{bevilacqua2012low} (PSNR (dB)).}
\label{suppfig1}
\end{figure}

We analyzed the advantages of our LIWT on the benchmark dataset for large-scale SR scenarios. We define scales larger than ×6 as large-scale. As shown in Table \ref{table13}, our method achieves optimal results for large-scale at ×6, ×8, and ×12. As shown in Figure \ref{suppfig1}, LIWT with RDN~\cite{zhang2018residual} as the encoder can even outperform other methods with SwinIR~\cite{liang2021swinir} as the encoder at ×8 SR on Set5~\cite{bevilacqua2012low}.

\begin{table}[h]
\begin{center}
\resizebox{0.9\columnwidth}{!}{
\begin{tabular}{|c|ccc|ccc|ccc|ccc|}
\hline
                          & \multicolumn{3}{c|}{Set5~\cite{bevilacqua2012low}}                                                                  & \multicolumn{3}{c|}{Set14~\cite{zeyde2012single}}                                                                 & \multicolumn{3}{c|}{B100~\cite{martin2001database}}                                                                  & \multicolumn{3}{c|}{Urban100~\cite{huang2015single}}                                                              \\ \cline{2-13} 
\multirow{-2}{*}{Methods} & $\times6$                           & $\times8$                           & $\times12$                          & $\times6$                           & $\times8$                           & $\times12$                          & $\times6$                           & $\times8$                           & $\times12$                          & $\times6$                           & $\times8$                           & $\times12$                          \\ \hline
RDN-Meta-SR~\cite{hu2019meta}               & 29.04                        & 26.96                        & -                            & 26.51                        & 24.97                        & -                            & 25.90                        & 24.83                        & -                            & 23.99                        & 22.59                        & -                            \\
RDN-LIIF~\cite{chen2021learning}                  & 29.15                        & 27.14                        & {\color[HTML]{0000FF} 24.86} & 26.64                        & 25.15                        & 23.24                        & 25.98                        & 24.91                        & 23.57                        & 24.20                        & 22.79                        & 21.15                        \\
RDN-UltraSR~\cite{xu2021ultrasr}               & {\color[HTML]{0000FF} 29.33} & 27.24                        & 24.81                        & {\color[HTML]{0000FF} 26.69} & {\color[HTML]{0000FF} 25.25} & {\color[HTML]{0000FF} 23.32} & {\color[HTML]{0000FF} 26.01} & {\color[HTML]{0000FF} 24.96} & {\color[HTML]{0000FF} 23.59} & {\color[HTML]{0000FF} 24.30} & 22.87                        & 21.20                        \\
RDN-IPE~\cite{liu2021enhancing}                   & 29.25                        & 27.22                        & -                            & 26.58                        & 25.09                        & -                            & 26.00                        & 24.93                        & -                            & 24.26                        & 22.87                        & -                            \\
RDN-LTE~\cite{lee2022local}                   & {\color[HTML]{343434} 29.32} & {\color[HTML]{0000FF} 27.26} & {\color[HTML]{343434} 24.79} & {\color[HTML]{343434} 26.71} & {\color[HTML]{343434} 25.16} & {\color[HTML]{343434} 23.31} & {\color[HTML]{0000FF} 26.01} & {\color[HTML]{343434} 24.95} & {\color[HTML]{343434} 23.6}  & {\color[HTML]{343434} 24.28} & {\color[HTML]{0000FF} 22.88} & {\color[HTML]{0000FF} 21.22} \\
RDN-LIWT (Ours)           & {\color[HTML]{FF0000} 29.45} & {\color[HTML]{FF0000} 27.38} & {\color[HTML]{FF0000} 25.00} & {\color[HTML]{FF0000} 26.80} & {\color[HTML]{FF0000} 25.30} & {\color[HTML]{FF0000} 23.36} & {\color[HTML]{FF0000} 26.05} & {\color[HTML]{FF0000} 24.99} & {\color[HTML]{FF0000} 23.63} & {\color[HTML]{FF0000} 24.41} & {\color[HTML]{FF0000} 23.01} & {\color[HTML]{FF0000} 21.33} \\ \hline
SwinIR-Meta-SR~\cite{hu2019meta}            & 29.09                        & 27.02                        & 24.82                        & 26.58                        & 25.09                        & 23.33                        & 25.94                        & 24.86                        & 23.59                        & 24.16                        & 22.75                        & 21.31                        \\
SwinIR-LIIF~\cite{chen2021learning}               & 29.46                        & {\color[HTML]{0000FF} 27.36} & {\color[HTML]{0000FF} 24.99} & 26.82                        & 25.34                        & 23.39                        & 26.07                        & 25.01                        & 23.64                        & 24.59                        & 23.14                        & 21.43                        \\
SwinIR-LTE~\cite{lee2022local}                & {\color[HTML]{0000FF} 29.50} & 27.35                        & {\color[HTML]{FF0000} 25.07} & {\color[HTML]{0000FF} 26.86} & {\color[HTML]{0000FF} 25.42} & {\color[HTML]{0000FF} 23.44} & {\color[HTML]{0000FF} 26.09} & {\color[HTML]{0000FF} 25.03} & {\color[HTML]{0000FF} 23.66} & {\color[HTML]{0000FF} 24.62} & {\color[HTML]{0000FF} 23.17} & {\color[HTML]{0000FF} 21.50} \\
SwinIR-LIWT (Ours)        & {\color[HTML]{FF0000} 29.60} & {\color[HTML]{FF0000} 27.51} & {\color[HTML]{FF0000} 25.07} & {\color[HTML]{FF0000} 26.90} & {\color[HTML]{FF0000} 25.43} & {\color[HTML]{FF0000} 23.48} & {\color[HTML]{FF0000} 26.13} & {\color[HTML]{FF0000} 25.06} & {\color[HTML]{FF0000} 23.69} & {\color[HTML]{FF0000} 24.71} & {\color[HTML]{FF0000} 23.24} & {\color[HTML]{FF0000} 21.57} \\ \hline
\end{tabular}
}
\end{center}
\caption{Comparison for scales at $\times2$, $\times3$, and $\times4$ on benchmark datasets (PSNR (dB)).}
\label{table13}
\end{table}

\begin{figure}[t!]
\centerline{\includegraphics[width=\columnwidth]{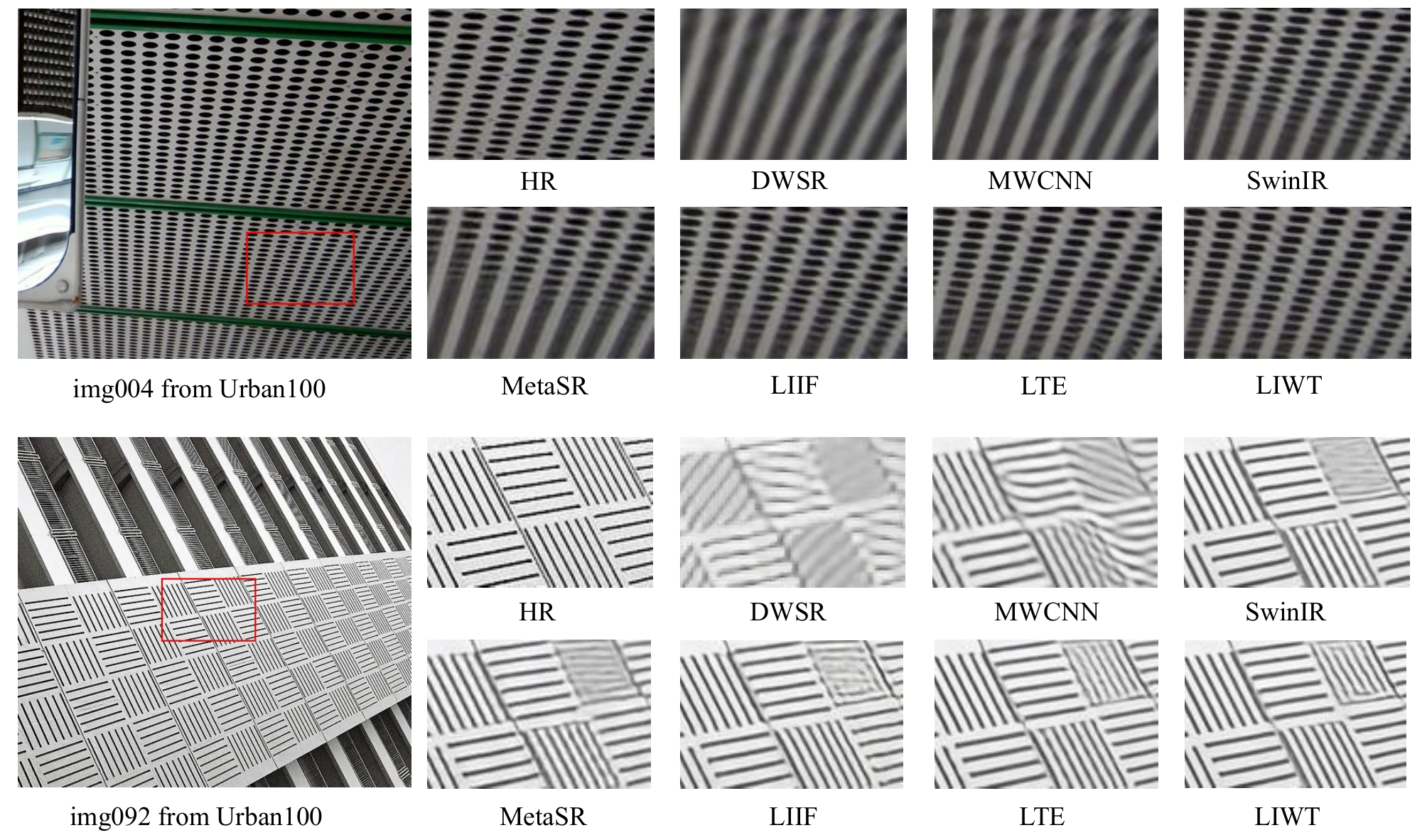}}
\caption{Visual comparisons for ×4 SR on Urban100~\cite{huang2015single}. Zoom in for best view.}
\label{suppfig2}
\end{figure}

\bibliography{egbib}
\end{document}